\documentclass[runningheads]{llncs}
\usepackage{graphicx}
\usepackage[misc]{ifsym}
\usepackage{multirow}
\usepackage{amssymb}

\newcommand{\tableImpute}{
    \begin{tabular}{|l|l|lr|lr|lr|}
        \hline
        Model & Metric & \multicolumn{2}{c|}{MYOCARDIAL} & \multicolumn{2}{c|}{~~~~~NHANES~~~~~} & \multicolumn{2}{c|}{~~~~~~COVID~~~~~~} \\
        \hline
        
        \multirow{4}{*}{OURS} & \multirow{2}{*}{BalACC} & $\mathbf{77.91\%}$ & \multirow{2}{*}{} & $\mathbf{64.17\%}$ & \multirow{2}{*}{} & $86.84\%$ & \multirow{2}{*}{} \\
        & & $\mathbf{\pm1.12\%}$ & & $\mathbf{\pm0.36\%}$ & & $\pm1.23\%$ & \\
        \cline{2-8}
        & \multirow{2}{*}{AUC} & $\mathbf{86.28\%}$ & \multirow{2}{*}{} & $\mathbf{69.92\%}$ & \multirow{2}{*}{} & $\mathbf{92.95\%}$ & \multirow{2}{*}{} \\
        & & $\mathbf{\pm0.42\%}$ & & $\mathbf{\pm0.56\%}$ & & $\mathbf{\pm0.93\%}$ & \\
        \hline
        
        \multirow{4}{*}{MEAN} & \multirow{2}{*}{BalACC} & $77.30\%$ & \multirow{2}{*}{$\equiv$} & $60.35\%$ & \multirow{2}{*}{$\bullet$} & $85.91\%$ & \multirow{2}{*}{$\bullet$} \\
        & & $\pm0.00\%$ & & $\pm0.00\%$ & & $\pm0.00\%$ & \\
        \cline{2-8}
        & \multirow{2}{*}{AUC} & $85.09\%$ & \multirow{2}{*}{$\bullet$} & $66.10\%$ & \multirow{2}{*}{$\bullet$} & $91.20\%$ & \multirow{2}{*}{$\bullet$} \\
        & & $\pm0.00\%$ & & $\pm0.00\%$ & & $\pm0.00\%$ & \\
        \hline

        \multirow{4}{*}{KNN} & \multirow{2}{*}{BalACC} & $68.83\%$ & \multirow{2}{*}{$\bullet$} & $63.00\%$ & \multirow{2}{*}{$\bullet$} & $\mathbf{88.08\%}$ & \multirow{2}{*}{$\circ$} \\
        & & $\pm0.00\%$ & & $\pm0.00\%$ & & $\mathbf{\pm0.00\%}$ & \\
        \cline{2-8}
        & \multirow{2}{*}{AUC} & $78.94\%$ & \multirow{2}{*}{$\bullet$} & $67.78\%$ & \multirow{2}{*}{$\bullet$} & $91.53\%$ & \multirow{2}{*}{$\bullet$} \\
        & & $\pm0.00\%$ & & $\pm0.00\%$ & & $\pm0.00\%$ & \\
        \hline

        \multirow{4}{*}{GAIN} & \multirow{2}{*}{BalACC} & $63.89\%$ & \multirow{2}{*}{$\bullet$} & $61.36\%$ & \multirow{2}{*}{$\bullet$} & $85.14\%$ & \multirow{2}{*}{$\bullet$} \\
        & & $\pm2.21\%$ & & $\pm0.53\%$ & & $\pm0.91\%$ & \\
        \cline{2-8}
        & \multirow{2}{*}{AUC} & $74.22\%$ & \multirow{2}{*}{$\bullet$} & $66.85\%$ & \multirow{2}{*}{$\bullet$} & $91.36\%$ & \multirow{2}{*}{$\bullet$} \\
        & & $\pm1.11\%$ & & $\pm0.40\%$ & & $\pm0.73\%$ & \\
        \hline

        \multirow{4}{*}{MICE} & \multirow{2}{*}{BalACC} & $76.55\%$ & \multirow{2}{*}{$\bullet$} & $61.70\%$ & \multirow{2}{*}{$\bullet$} & $87.98\%$ & \multirow{2}{*}{$\circ$} \\
        & & $\pm0.00\%$ & & $\pm0.00\%$ & & $\pm0.00\%$ & \\
        \cline{2-8}
        & \multirow{2}{*}{AUC} & $81.39\%$ & \multirow{2}{*}{$\bullet$} & $67.30\%$ & \multirow{2}{*}{$\bullet$} & $92.43\%$ & \multirow{2}{*}{$\equiv$} \\
        & & $\pm0.00\%$ & & $\pm0.00\%$ & & $\pm0.00\%$ & \\
        \hline

        \multirow{4}{*}{MISSFOREST} & \multirow{2}{*}{BalACC} & $73.00\%$ & \multirow{2}{*}{$\bullet$} & $61.40\%$ & \multirow{2}{*}{$\bullet$} & $85.15\%$ & \multirow{2}{*}{$\bullet$} \\
        & & $\pm0.87\%$ & & $\pm1.03\%$ & & $\pm1.67\%$ & \\
        \cline{2-8}
        & \multirow{2}{*}{AUC} & $80.82\%$ & \multirow{2}{*}{$\bullet$} & $66.48\%$ & \multirow{2}{*}{$\bullet$} & $91.30\%$ & \multirow{2}{*}{$\bullet$} \\
        & & $\pm1.60\%$ & & $\pm0.90\%$ & & $\pm1.20\%$ & \\
        \hline

        \multirow{4}{*}{SOFTIMPUTE} & \multirow{2}{*}{BalACC} & $77.24\%$ & \multirow{2}{*}{$\equiv$} & $61.70\%$ & \multirow{2}{*}{$\bullet$} & $84.48\%$ & \multirow{2}{*}{$\bullet$} \\
        & & $\pm0.99\%$ & & $\pm0.93\%$ & & $\pm0.78\%$ & \\
        \cline{2-8}
        & \multirow{2}{*}{AUC} & $84.88\%$ & \multirow{2}{*}{$\bullet$} & $66.93\%$ & \multirow{2}{*}{$\bullet$} & $91.12\%$ & \multirow{2}{*}{$\bullet$} \\
        & & $\pm0.77\%$ & & $\pm1.08\%$ & & $\pm0.85\%$ & \\
        \hline

        \multirow{4}{*}{SINKHORN} & \multirow{2}{*}{BalACC} & $75.66\%$ & \multirow{2}{*}{$\bullet$} & $60.77\%$ & \multirow{2}{*}{$\bullet$} & $86.82\%$ & \multirow{2}{*}{$\equiv$} \\
        & & $\pm1.22\%$ & & $\pm0.98\%$ & & $\pm1.49\%$ & \\
        \cline{2-8}
        & \multirow{2}{*}{AUC} & $83.26\%$ & \multirow{2}{*}{$\bullet$} & $65.42\%$ & \multirow{2}{*}{$\bullet$} & $91.48\%$ & \multirow{2}{*}{$\bullet$} \\
        & & $\pm1.01\%$ & & $\pm1.18\%$ & & $\pm1.13\%$ & \\
        \hline

        \multirow{4}{*}{MIDA} & \multirow{2}{*}{BalACC} & $75.09\%$ & \multirow{2}{*}{$\bullet$} & $62.15\%$ & \multirow{2}{*}{$\bullet$} & $85.55\%$ & \multirow{2}{*}{$\bullet$} \\
        & & $\pm0.70\%$ & & $\pm1.26\%$ & & $\pm1.12\%$ & \\
        \cline{2-8}
        & \multirow{2}{*}{AUC} & $82.87\%$ & \multirow{2}{*}{$\bullet$} & $66.91\%$ & \multirow{2}{*}{$\bullet$} & $91.67\%$ & \multirow{2}{*}{$\bullet$} \\
        & & $\pm0.78\%$ & & $\pm1.30\%$ & & $\pm0.62\%$ & \\
        \hline
    \end{tabular}
}
\usepackage{hyperref}
\hypersetup{pdfauthor={Thomas Ranvier}}

\begin{document}
\title{Autoencoder-based Attribute Noise Handling Method for Medical Data}

\toctitle{Autoencoder-based Attribute Noise Handling Method for Medical Data}
\tocauthor{Thomas~Ranvier}

% authors
\author{Thomas Ranvier \Letter\inst{1,2}\orcidID{\href{https://orcid.org/0000-0001-9250-9530}{0000-0001-9250-9530}} \and
Haytham Elgazel\inst{1,3} \and
Emmanuel Coquery\inst{1,4} \and
Khalid Benabdeslem\inst{1,5}
}
\authorrunning{T. Ranvier et al.}
\institute{
Univ Lyon, UCBL, CNRS, INSA Lyon, LIRIS, UMR5205\\
43 bd du 11 Novembre 1918, 69622 Villeurbanne, France \and
\email{thomas.ranvier@univ-lyon1.fr} \and
\email{haytham.elghazel@univ-lyon1.fr} \and
\email{emmanuel.coquery@liris.cnrs.fr} \and
\email{khalid.benabdeslem@univ-lyon1.fr}
}
\maketitle
\begin{abstract}%150-250
Medical datasets are particularly subject to attribute noise, that is, missing and erroneous values.
Attribute noise is known to be largely detrimental to learning performances.
To maximize future learning performances it is primordial to deal with attribute noise before any inference.
We propose a simple autoencoder-based preprocessing method that can correct mixed-type tabular data corrupted by attribute noise.
No other method currently exists to handle attribute noise in tabular data.
We experimentally demonstrate that our method outperforms both state-of-the-art imputation methods and noise correction methods on several real-world medical datasets.

\end{abstract}
\section{Introduction}

% noise in real-world and especially medical data and importance for ML
Medical studies are particularly subject to outliers, erroneous, meaningless, or missing values.
In most real-life studies, not solely limited to the medical field, the problem of incomplete data and erroneous data is unavoidable.
Those corruptions can occur at any data collection step.
They can be a natural part of the data (patient noncompliance, irrelevant measurement, etc.) or appear from corruption during a later data manipulation phase \cite{yang_dealing_2004}.
Regardless of their origin, those corruptions are referred as ``noise'' in the following work.
Noise negatively impacts the interpretation of the data, be it for a manual data analysis or training an inference model on the data.
The goal of a machine learning model is to learn inferences and generalizations from training data and use the acquired knowledge to perform predictions on unseen test data later on.
Thus, the quality of training data on which a model is based is of critical importance, the less noisy the data is, the better results we can expect from the model.

% 2 types of noise
Noise can be divided into two categories, namely class noise and attribute noise \cite{zhu_class_2004}.
Class noise corresponds to noise in the labels, e.g. when data points are labeled with the wrong class, etc.
Attribute noise on the other hand corresponds to erroneous and missing values in the attribute data, that is, the features of the instances.
Attribute noise tends to occur more often than class noise in real-world data \cite{yang_dealing_2004}\cite{zhu_class_2004}\cite{van_hulse_pairwise_2007}.
Despite this fact, compared to class noise, very limited attention has been given to attribute noise \cite{zhu_class_2004}.
In real-world medical data, the probability of mislabeled data in a survival outcome context is quite low, we focused our work on attribute noise to maximize prediction performance while trying to compensate for a lack of appropriate methods within the literature.

% missing values methods
The problem of imputing missing values has been vastly addressed in the literature, one can choose from many imputation methods to complete its data depending on its specific needs \cite{stef_flexible_2018}.
Imputation methods only address part of the attribute noise problem, they can handle missing values but do not handle erroneous values, which can be highly detrimental to imputation results.
Those methods have been widely researched, but methods able to deal with erroneous values have been less researched and can be considered incomplete at the moment \cite{yang_dealing_2004}.

% noise correction methods
Handling erroneous values can be done in three main ways: using robust learners that can learn directly from noisy data and naturally compensate or partially ignore the noise, filtering methods that remove data points that are classified as noisy, and polishing methods that aim to correct noisy instances.
Robust learners are models that are less sensitive to noise in the data than classic models but they present several disadvantages \cite{van_hulse_pairwise_2007}.
They usually have limited learning potential compared to other learners.
Using robust learners is not useful if we aim to perform anything else than the task the learner will solve.
Filtering methods aim to detect which instances are noisy to delete them from the training set \cite{zhu_class_2004}\cite{van_hulse_pairwise_2007}.
By training a learner on this cleaned set it can learn inferences without being disturbed by erroneous values and outliers which eventually leads to better prediction performances on test data.
The third way to deal with erroneous values is the polishing method \cite{teng_polishing_2004}, which corrects instances detected as noisy.
Such a method can correct erroneous values on small datasets but lacks scalability for larger datasets containing more features \cite{van_hulse_pairwise_2007}.
Those three methods are able to deal with erroneous values and outliers but are not able to deal with incomplete data, they only address part of the attribute noise problem.

% mini conclusion on those methods
At the moment the only way to handle attribute noise in its entirety is to use a combination of an imputation method followed by a noise correction method, to the best of our knowledge the literature lacks a method that would be able to perform both those tasks at once.
Real-world data and especially medical data are subject to attribute noise in its entirety, it is important to conceive an approach able to handle the totality of attribute noise and not just subpart of it.

% contributions
In this paper, we propose a preprocessing method based on autoencoders that deals with attribute noise in its entirety in real-world tabular and mixed-type medical data.
Our method is able to learn from incomplete and noisy data to produce a corrected version of the dataset.
It does not require any complete instance in the dataset and can truly handle attribute noise by performing both missing values completion and correction of erroneous values at the same time.
We conduct extensive experiments on an imputation task on real-world medical data to compare our method to other state-of-the-art methods and obtain competitive and even significantly better results on classification tasks performed on the corrected data.
We extend our experiments to show that our method can both complete missing data while correcting erroneous values, which further improves the obtained results.

The complete source code used to conduct the experiments is available at the following github repository\footnote{\url{https://github.com/ThomasRanvier/Autoencoder-based_Attribute_Noise_Handling_Method_for_Medical_Data}}.

% plan
The rest of the paper is organized as follows: we first present related work of data imputation and noise correction in both tabular and image data, especially in the medical field, in section \ref{related_work}.
Then, we present and explain our proposed approach in section \ref{our_approach}.
Section \ref{results} shows our experimental results compared to both data imputation and noise correction state-of-the-art methods.
Finally, we conclude with a summary of our contributions.

\section{Related work \label{related_work}}

Denoising is vastly researched in the image field, in the image medical domain it is easy to find recent reviews and methods to correct medical images \cite{mohd_sagheer_review_2020}.
Correction of tabular data on the other hand is less researched, only the imputation part seems to attract lots of attention.
In this paper, we are especially focused on methods that can be applied to mixed-type tabular data.

Recently lots of autoencoder-based imputation methods have been researched \cite{pereira_reviewing_2020}.
An autoencoder is a machine learning algorithm that takes an input $x \in \mathbb{R}^{d}$ with $d$ the number of features and learns an intermediate representation of the data noted $z \in \mathbb{R}^{h}$ with $h$ the size of the newly constructed latent space.
Then, from the intermediate representation $z$ the model reconstructs the original data $x$, we note the model output $\hat{x}$.
During its training, the reconstruction error between $x$ and $\hat{x}$ is minimized.

One of those autoencoder-based imputation methods is MIDA: Multiple Imputation using Denoising Autoencoders, introduced in 2018 \cite{gondara_mida_2018}.
Unlike most autoencoder-based methods which are usually applied to images, MIDA has been successfully applied to tabular data.
This imputation method learns from a complete training dataset and can then be applied to unseen incomplete test data to impute the missing values.
The authors assume that in order to learn how to impute missing values MIDA must learn from complete data.
However, in this paper our experimental protocol does not provide a clean dataset to train on, therefore we show that MIDA obtains satisfactory results when properly parameterized, even when learning on incomplete data.

We want to show that autoencoders can not only be used to impute missing values, but also to correct erroneous values that are part of the observed values.
It is easier to correct erroneous values in an image than in tabular data, since in images pixels in a close neighborhood are related to each other, which might not be true for arbitrarily ordered features in tabular data.
As stated earlier correction of images is a very active research domain.
Recently, Ulyanov et al. introduced a new approach called ``Deep Image Prior \cite{ulyanov_deep_2020}.''
That innovative approach uses autoencoders to restore images but does not use the original data $X$ as model input, instead, the autoencoder is given pure noise as input and is trained to reconstruct the original corrupted data $X$ from the noise.
In that way, the model is no longer considered an autoencoder but a generative model, however, in practice the model keeps the same architecture.
Therefore, the only information required to correct the input image is already contained in the image itself.
By stopping the training before complete convergence it is possible to obtain a cleaner image than the original corrupted image.
Ulyanov et al. showed that their approach outperforms other state-of-the-art methods on a large span of different tasks.

In this paper, we aim to conceive a method that would be able to correct mixed-type tabular data, we aim to use the lessons from \cite{gondara_mida_2018} and \cite{ulyanov_deep_2020} to conceive a method able to handle attribute noise as a whole as a preprocessing method.

\section{A Method to Truly Handle Attribute Noise \label{our_approach}}

\begin{figure}[!ht]
    \centering
    \includegraphics[width=.7\textwidth]{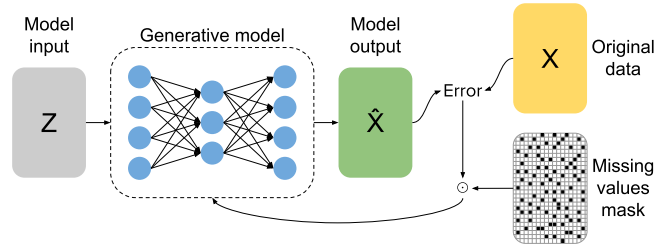}
    \caption{The model parameters are trained so that the model learns to reconstruct the original data. The training must be stopped at the right moment for the reconstruction $\hat{X}$ to be cleaner than the original data $X$.}
    \label{simple_schema}
\end{figure}

Our method is based on a deep neural architecture that is trained to reconstruct the original data from a random noise input.
We note the original data with its attribute noise $X \in \mathbb{R}^{n \times d}$ with $n$ the number of instances in the dataset and $d$ the number of features.
We note the deep generative model $\hat{X} = f_\Theta(\cdot)$ with $\Theta$ the model parameters that are learned during training and $\hat{X}$ is the model output, in our case the model output is a reconstruction of $X$.
The input of the model is noted $Z \in \mathbb{R}^{n \times d}$ and has the same dimension as $X$, which keeps our model a kind of autoencoder.
The model is trained to reconstruct $X$ using the following loss term: $L(X, \hat{X}) = ||(\hat{X} - X) \odot M||^2$, where $\odot$ is the Hadamard product and $M \in \mathbb{R}^{n \times d}$ is a binary mask that stores the locations of missing values in $X$, $M_{ij} = 1$ if $X_{ij}$ is observed and $M_{ij} = 0$ if $X_{ij}$ is missing.
By applying the mask $M$ to the loss ensure that the loss is only computed on observed values, in this way the reconstruction $\hat{X}$ will fit the observed values in $X$, while missing values will naturally converge to values that are statistically consistent given the learned data distribution.
Figure \ref{simple_schema} shows how the model is fitted to the original data $X$ with the application of the binary mask $M$ during training. 

To determine the right step at which to stop training we define a stopping condition based on the evolution of a given metric.
In a supervised setting, for example, we regularly compute the AUC on a prediction task performed on the reconstructed data $\hat{X}$, which gives us the evolution of the quality of the reconstruction.
We stop the training when the AUC degrades for a set number of iterations, then we obtain a reconstruction with consistent imputations and noise correction, which provides better data quality than the original data.
If no supervision is possible the only metric that can be used to determine when to stop training is the loss value, which gives correct results but is quite limited since it is harder to determine when overfitting starts.

In practice, we set the input of the model either as random noise or as the original data depending on the obtained results.
We note that on datasets containing large amounts of features, using 1D convolutions instead of classic fully-connected layers tends to give better results, it helps our method to scale on datasets with large amounts of data and features.

What makes this method different from most other autoencoder-based methods and able to handle attribute noise as a whole is that we determine an early-stop condition to stop training at the proper moment.
Stopping at the right moment allows the method to reach a point where the reconstructed data $\hat{X}$ contains less noise than the original data $X$ while containing all important information from $X$.

%We made an automatic method that makes use of the hyperparameters optimization process from \cite{optuna} to determine the best parameters for a given dataset and returns the best found correction.
%That method has the advantage that it only requires to find a coherent epoch number that depends on how much the user wants the method to loop over the data at maximum.
%Other than that all hyperparameters are automatically tuned.
%The main disadvantage of the automatic method is that it takes a long amount of time to tune the parameters and return the best correction.
%It should only be used by users that don't have sufficient knowledge about machine learning and/or not enough time to manually find good parameters for our method.

\section{Experimental Results \label{results}}

\subsection{Used Datasets}

We ran our experiments on three real-life medical mixed-type tabular datasets naturally containing missing values.
We evaluated our method and compared our results to other state-of-the-art methods on those medical datasets.

\begin{itemize}
    \item NHANES, US National Health and Nutrition Examination Surveys: Those are surveys conducted periodically by the US NHCS to assess the health and nutritional status of the US population \cite{barnard_applications_1999}.
    We used data from studies spanning from $2000$ to $2008$, with $95$ features and about $33\%$ missing values.
    We selected the ``diabete'' feature as a class and randomly selected $1000$ samples from both outcomes to evaluate the quality of the data correction on a classification task on this class.
    \item COVID19: This dataset was publicly released with the paper \cite{yan_interpretable_2020}, it contains medical information collected between in early 2020 on pregnant and breastfeeding women.
    We based our data preprocessing on the one realized in the original paper, we selected only the measurements from the last medical appointment for each patient.
    After preprocessing, we obtain a dataset composed of $361$ patients with $76$ features, with about $20\%$ missing data.
    We evaluate the quality of the data correction on a classification task on the survival outcome, $195$ patients have survived and $166$ are deceased.
    \item Myocardial infarction complications: This medical dataset is available on the UCI machine learning repository, it was publicly released with the paper \cite{golovenkin_trajectories_2020}.
    It is composed of $1700$ patients with $107$ features, with about $5\%$ missing values.
    We evaluate the quality of the data correction on a classification task on the survival outcome, $1429$ patients have survived and $271$ are deceased.
\end{itemize}

\subsection{Used Metrics}

We evaluate the quality of the obtained correction on classification tasks.
As can be seen from the previous section, the medical datasets we used are not all balanced, the Myocardial dataset is especially imbalanced.
In such a context it is important to choose metrics that are not sensitive to imbalance.

In a medical context where we aim to predict the outcome between sane and sick, it is extremely important not to classify sick patients as sane since it would be very detrimental for them not to get an appropriate medical response.
%On the other hand, classifying a patient as sick when it is not the case is less detrimental since it would most likely result in more medical investigations for this patient but would not put its life at risk.
In machine learning terms we are in cases where false positives on the negative class would be less detrimental than false negatives, therefore we should aim to minimize false negatives.

Appropriate metrics, in this case, are the AUC: Area Under the Receiver Operating Characteristic (ROC) Curve, and the balanced accuracy.
The AUC corresponds to the area under the ROC curve obtained by plotting the true positive rate (recall) against the false positive rate (1-specificity).
An AUC score of $1$ would mean that the classifier gives true positives $100\%$ of the time, whereas a value of $0.5$ means that the classifier is no better than a random prediction.
The balanced accuracy is defined as the average of recall obtained on each class, which is simply the average of the true positive rate between all the classes.

\subsection{Experimental Protocol}

% introduction
Our experiments aim to compare our method to other state-of-the-art methods for both imputation and noise correction tasks.
All our experiments are evaluated using the balanced accuracy and the AUC metrics.
We repeated each experiment 10 times with 10 different stochastic seeds to set up the random state of non-deterministic methods.
For each experiment we compare the performances of each method to ours using t-tests.
We use the results from those statistical tests to determine if our method is significantly better, even, or significantly worse than each other method, based on a $p$-value set at $0.05$.

% medical data imputation
We first evaluate our method on an imputation task on the three medical datasets previously described.
Each of those datasets is missing part of its data, we compare the quality of the data imputation by training a decision tree on a classification task on each dataset after imputation.

% data imputation on noisy data
The capacity of our method to impute missing values on incomplete and noisy data is assessed by introducing artificial noise in the datasets. Noise is artificially added to the data by randomly replacing attribute values with a random number at a certain rate, as described by Zhu et al. \cite{zhu_class_2004}.
We compare the results at the following noise rates: $0/5/10/15/20/40/60\%$.

% data imputation + noise correction
Finally, to assess the effectiveness of our method to both complete missing values while correcting erroneous values, we introduce artificial noise in the naturally incomplete datasets. We apply our method and compare its results to those obtained by a sequential execution (\textit{i.e.} pipeline) of an imputation method followed by a noise correction method.

% imputation methods
The results from our method are compared to those of other state-of-the-art imputation methods:
\begin{itemize}
    \item MEAN, MEDIAN and KNN: We used the ``SimpleImputer'' and ``KNNImputer'' classes from the python library ``scikit-learn''\footnote{\url{https://scikit-learn.org}}.
    \item MICE: Multivariate Imputation by Chained Equations has been introduced in 2011 in \cite{buuren_mice_2011}.
    This is a very popular method of imputation because it provides fast, robust, and good results in most cases. 
    We used the implementation from the experimental ``IterativeImputer'' class from ``scikit-learn''.
    \item GAIN: Generative Adversarial Imputation Nets, introduced recently in \cite{yoon_gain_2018}, two models are trained in an adversarial manner to achieve good imputation.
    We used the implementation from the original authors\footnote{\url{https://github.com/jsyoon0823/GAIN}}.
    \item SINKHORN: An optimal transport based method for data imputation introduced in \cite{muzellec_missing_2020}
    We used the implementation from the original authors\footnote{\url{https://github.com/BorisMuzellec/MissingDataOT}}.
    \item SOFTIMPUTE: The SOFTIMPUTE algorithm has been proposed in 2010 \cite{mazumder_spectral_2010}, it iteratively imputes missing values using an SVD.
    We used the public re-implementation by Travis Brady of the Mazumder and Hastie's package\footnote{\url{https://github.com/travisbrady/py-soft-impute}}.
    \item MISSFOREST: An iterative imputation method based on random forests introduced in 2012 in \cite{stekhoven_missforestnon-parametric_2012}.
    We used the ``MissForest'' class from the python library ``missingpy''\footnote{\url{https://pypi.org/project/missingpy/}}.
    \item MIDA: Multiple Imputation Using Denoising Autoencoders has been recently proposed in \cite{gondara_mida_2018}.
    We implemented MIDA using the author description from the original paper and the code template supplied in this public gist \footnote{\url{https://gist.github.com/lgondara/18387c5f4d745673e9ca8e23f3d7ebd3}}
\end{itemize}

% noise correction methods
The following noise correction methods are used as comparison:
\begin{itemize}
    \item SFIL: Standard Filtering, which we implemented such as described in \cite{teng_polishing_2004}.
    \item SPOL: Standard Polishing, which we also implemented such as described in \cite{teng_polishing_2004}.
    \item PFIL, PPOL: Improved versions of SFIL and SPOL where the noisy instances to filter or polish are identified using the noise detection method Panda \cite{van_hulse_pairwise_2007}.
\end{itemize}

\subsection{Results}

In this subsection, we present and analyze the most important comparative results between our proposed method and state-of-the-art methods.
The entire experimental results with statistical significance can be found in our code.

%\vspace{-5mm}
\subsubsection{Imputation on Incomplete Medical Data}

With our first experiment, we show that our method can impute missing values in real-world medical datasets.

\begin{table}[!ht]
    \centering
    \caption{Comparative study between our method and other methods. BalACC corresponds to the balanced accuracy, AUC is the area under the ROC curve. Our method is compared to each other using t-tests with a $p$-value of $0.05$, when our method is significantly better it is indicated by $\bullet$, even by $\equiv$, and significantly worse by $\circ$.}
    \label{tab_expe_1}
    \tableImpute
\end{table}

Table \ref{tab_expe_1} shows the results on the three real-world medical datasets.
We can see that our method obtains very competitive results on all datasets.
We obtain significantly better results than other state-of-the-art methods in most cases for both metrics.
The only cases in which our method performs significantly worse are against KNN and MICE on COVID data on the balanced accuracy metric.
This shows that our method is able to impute missing values on incomplete real-world medical mixed-type tabular data with results as good as other state-of-the-art imputation methods and even better in most cases.

%\vspace{-5mm}
\subsubsection{Imputation on Incomplete and Noisy Medical Data}

Our second experiment shows that our method can impute missing values in real-world medical datasets in a noisy context.
We artificially add noise to the data at various rates: $0/5/10/15/20/40/60\%$, and evaluate each imputation method at each noise level.

\begin{figure}[!ht]
    \centering
    %\resizebox{\textwidth}{!}{\plotExpeTwo}
    \includegraphics[width=\textwidth]{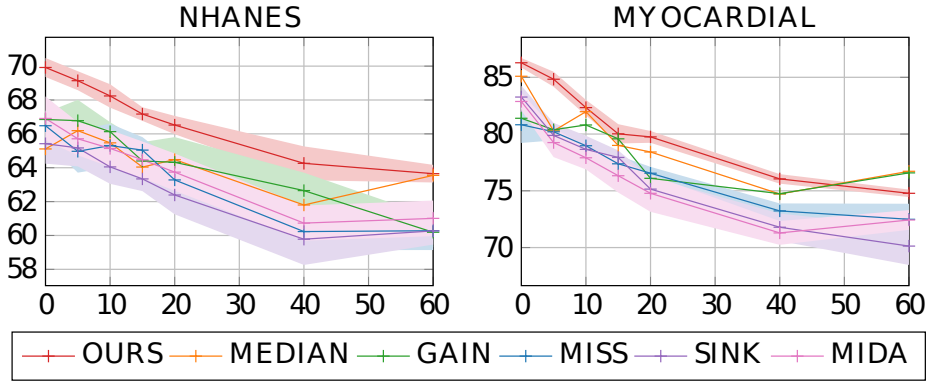}
    \caption{AUC results on imputation on incomplete and noisy medical data}
    \label{plot_expe_2}
\end{figure}

Figure \ref{plot_expe_2} shows AUC results obtained on NHANES and MYOCARDIAL data at each noise rate against several imputation methods.
In both cases, we note that our method globally obtains significantly better results than other methods.
The performance of all methods drops when the noise level increases, which is expected.
On NHANES data our method performs largely better than others until a noise rate of $60\%$ where the MEDIAN imputation gets similar results to ours.
This can probably be explained by the fact that with a noise level that high it is nearly impossible to impute coherent values other than the median or mean value for each feature.
We can observe the same pattern on MYOCARDIAL data, with the difference that GAIN seems to have learned how to adapt to such an amount of noise in this case.
Those results show that on low to high noise rates, our method can impute missing values while correcting erroneous values.
It provides better data correction than most other methods.
At extreme noise rates naive methods might provide better results.

%\vspace{-5mm}
\subsubsection{Comparison with the Combination of Imputation and Noise Correction Methods}

The last experiment compares our method results to those obtained from the combination of an imputation method followed by a noise correction method.
We chose MICE as the state-of-the-art imputation method since it obtains competitive results against ours in a not noisy context.
We then apply the four noise correction methods SFIL, PFIL, SPOL, and PPOL.

\begin{figure}[!ht]
    \centering
    %\resizebox{\textwidth}{!}{\plotExpeThree}
    \includegraphics[width=\textwidth]{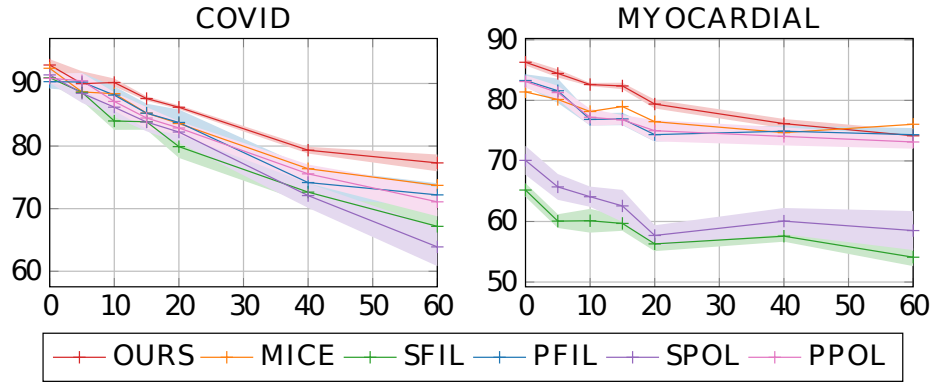}
    \caption{AUC results on combination of imputation and noise correction}
    \label{plot_expe_3}
\end{figure}

Figure \ref{plot_expe_3} show AUC results obtained on COVID and MYOCARDIAL data at each noise rate.
We note that SFIL and SPOL perform worse than the Panda alternative of both those methods at all noise rates.
We also note that for both datasets the other state-of-the-art noise correction methods give very poor results as soon as the noise level reaches more than $5\%$, at higher noise rates the data quality is better before noise correction than before.
For COVID data all methods yield similar results at low noise levels, with our method on top with a very slight advantage.
At high rates, however, our method gives very significantly better results than all other methods.
For MYOCARDIAL data the opposite pattern can be observed, our method gives significantly better results up until a noise rate of $40\%$, after which MICE imputation is slightly better.
This experiment completes the conclusions drawn from the second experiment, our method provides very good data correction, up until the noise rate becomes too extreme, at that point, simpler methods achieve slightly better results.
The fact that the opposite is observed on COVID data is probably due to a remarkable original data quality, which would explain why our method becomes significantly better only at higher noise levels.

\section{Conclusion}

Handling attribute noise means imputing missing values while correcting erroneous values and outliers.
This phenomenon is of critical importance in medical data, where attribute noise is especially present and detrimental to analysis and learning tasks on the data.
No method in the literature is capable of handling attribute noise in its entirety in mixed-type tabular data.
Many methods exist to impute missing values while other methods can correct erroneous values.

In this paper, we propose an autoencoder-based preprocessing approach to truly handle attribute noise.
Our method imputes missing values while correcting erroneous values without requiring any complete or clean instance in the dataset to correct.
Our experiments show that our method competes against and even outperforms other imputation methods on real-world medical mixed-type tabular data. Our method is less sensitive to noise on an imputation task. 

Finally, as autoencoder approaches are amenable to an empirical tuning phase, we plan to implement in the future an algorithm able to automatically define an adapted architecture depending on the  dataset dimensions.

\bibliographystyle{splncs04}
\bibliography{main}

\section*{Acknowledgments}

This research is supported by the European Union's Horizon 2020 research and innovation program under grant agreement No 875171, project QUALITOP (Monitoring multidimensional aspects of QUAlity of Life after cancer ImmunoTherapy - an Open smart digital Platform for personalized prevention and patient management).

\end{document}